\documentclass[10pt,twocolumn,letterpaper]{article}

\usepackage{wacv}
\usepackage{times}
\usepackage{epsfig}
\usepackage{graphicx}
\usepackage{amsmath}
\usepackage{amssymb}

\usepackage{times}
\usepackage{soul}
\usepackage{url}
\usepackage[utf8]{inputenc}
\usepackage[small]{caption}
\usepackage{booktabs}

\usepackage{algorithmic}
\usepackage[ruled,vlined]{algorithm2e}
\usepackage{bm}
\usepackage{url}
\usepackage{multirow}
\usepackage{arydshln}
\usepackage{pbox}
\usepackage{float}
\restylefloat{table} 
\usepackage{gensymb}
\usepackage{textcomp}
\usepackage{times}
\usepackage{amsmath}
\usepackage{amssymb}
\usepackage{float}
\usepackage{enumitem}
\DeclareMathOperator*{\argmax}{arg\,max}

\usepackage{color}
\usepackage{subfigure}
\usepackage{listings}

\usepackage{multicol}

\graphicspath{{./figures/}}

\definecolor{mygreen}{rgb}{0,0.6,0}
\definecolor{mygray}{rgb}{0.5,0.5,0.5}
\definecolor{mymauve}{rgb}{0.58,0,0.82}

\usepackage{xcolor}

\allowdisplaybreaks

\def\wacvPaperID{222} 

\wacvfinalcopy 
\ifwacvfinal
\def\assignedStartPage{9876} 
\fi

\ifwacvfinal
\usepackage[breaklinks=true,bookmarks=false]{hyperref}
\else
\usepackage[pagebackref=true,breaklinks=true,colorlinks,bookmarks=false]{hyperref}
\fi

\ifwacvfinal
\else
\fi

\begin{document}

\title{PROVES: Establishing Image Provenance using Semantic Signatures}

\author{Mingyang Xie$^1$, Manav Kulshrestha$^2$, Shaojie Wang$^1$, Jinghan Yang$^1$\\ Ayan Chakrabarti$^1$, Ning Zhang$^1$, and Yevgeniy Vorobeychik$^1$
\\\normalsize{$^1$Computer Science \& Engineering, Washington University in St.~Louis}\\ \normalsize{$^2$Computer Science, University of Massachusetts at Amherst}
}

\maketitle

\begin{abstract}
Modern AI tools, such as generative adversarial networks, have transformed our ability to create and modify visual data with photorealistic results.
However, one of the deleterious side-effects of these advances is the emergence of nefarious uses in manipulating information in visual data, such as through the use of deep fakes.
We propose a novel architecture for preserving the provenance of semantic information in images to make them less susceptible to deep fake attacks.
Our architecture includes semantic signing and verification steps.
We apply this architecture to verifying two types of semantic information: individual identities (faces) and whether the photo was taken indoors or outdoors.
Verification accounts for a collection of common image transformation, such as translation, scaling, cropping, and small rotations, and rejects adversarial transformations, such as adversarially perturbed or, in the case of face verification, swapped faces.
Experiments demonstrate that in the case of provenance of faces in an image, our approach is
robust to black-box adversarial transformations (which are rejected) as well as benign transformations (which are accepted), with few false negatives and false positives.
Background verification, on the other hand, is susceptible to black-box adversarial examples, but becomes significantly more robust after adversarial training.
\end{abstract}



\section{Introduction}\label{sec:intro}

Modern machine learning (ML)-enabled image synthesis tools~\cite{brock2018large,Dekel_2018_CVPR,pix2pix2017,suwajanakorn2017synthesizing,thies2018headon,Zhan_2019_CVPR} allow the editing and manipulation of images to create photorealistic results. 
While these tools have significant utility in graphic design and entertainment, they also make it challenging to combat the spread of  misinformation. 
\begin{figure}[h]
\centering
\begin{tabular}{cc}
\includegraphics[width=1.4in]{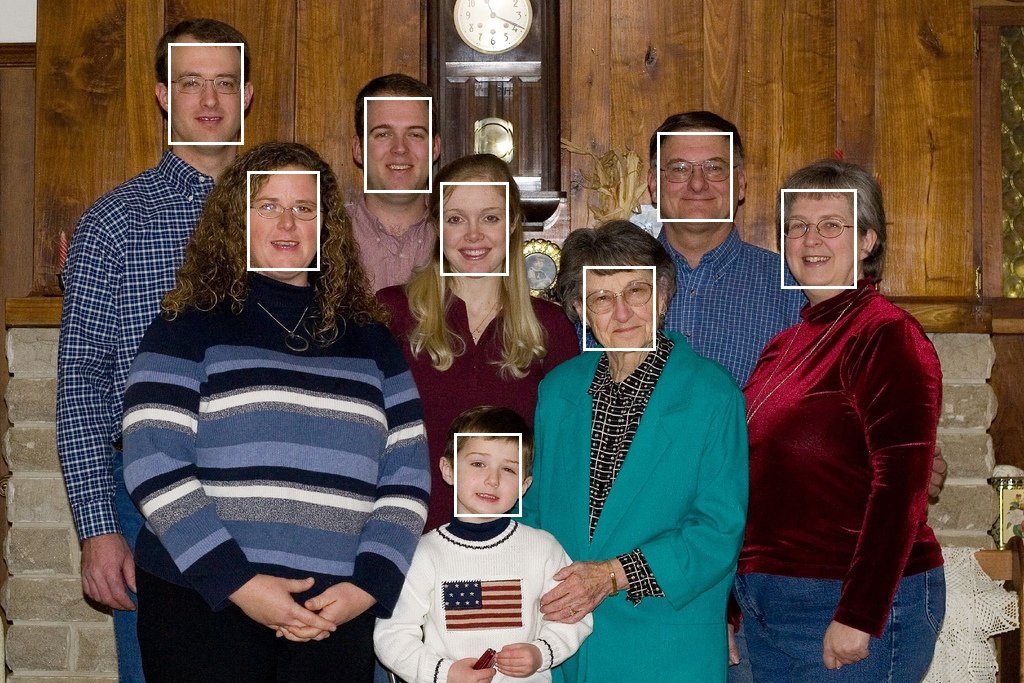} &
\includegraphics[width=1.4in]{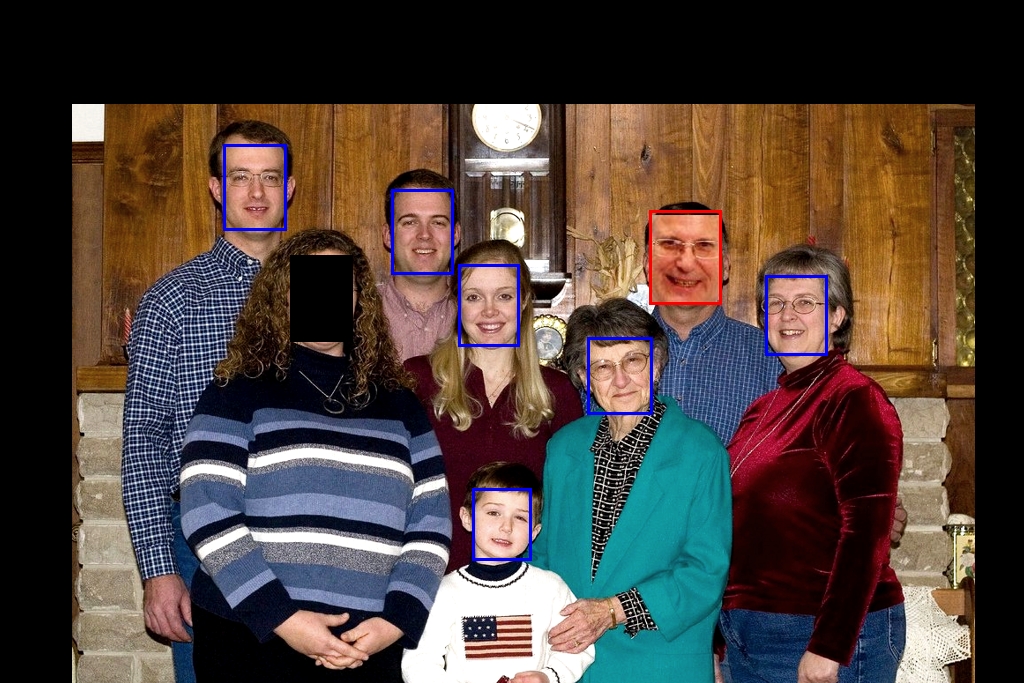}
\end{tabular}
  \caption{Illustration of \textsc{PROVES} verification. Left: original image. Right: image after translation, malicious face swapping and face occlusion. All unperturbed faces are correctly identified (blue bounding boxes). The occluded face is ignored, 
  and the swapped face is marked by a red bounding box.}
 \label{F:example}
\end{figure}
Common mitigation approaches attempt to detect fake images~\cite{Bappy17,Bayar16,mo2018fake,roessler2019faceforensicspp,Huh2018,Zhou2018}, but such approaches typically require the knowledge of the technique used to generate these, creating an arms race between fake image synthesis and detection methods.

We propose an alternative approach for ensuring image provenance that leverages 
cryptographic signatures produced by a trusted entity (e.g., a camera or a reputable photographer) at the time of capture to enable verification of the \emph{semantic content} of images as authentic. 
A trivial approach for this would be to establish a registry of public keys for trusted sources, have trusted sources simply sign the raw image pixel data with their private key and attach the signature to the image file, and verify the signatures of all images being uploaded or distributed to end-users.

However, requiring that the raw pixel data remain unmodified from the time of signing to the time an image is presented to the user is highly restrictive. Content creators often wish to edit an image in good faith for style, resolution, etc, without changing its semantic content.
To address this need, we propose a system that enables certification and verification of \emph{semantic information} in images as having been unmodified from the time the image was signed, while still allowing common editing operations.
We instantiate this system to certify two kinds of semantic information: 1) the identity and relative location of faces in an image and 2) whether the background represents a photo taken indoors or outdoors.
In order to make more general use of the framework, one only needs to define a sufficient collection of \emph{semantic categories}, beyond these two, as well as the relevant spatial relationships among them.
For example, just as we show how to offer provenance of the relative spatial relationships among individuals in an image, we can similarly certify spatial relationships among arbitrary objects in an image.
Similarly, just as we robustly detect and verify faces, we can do so for other entities in images.

Our system is architected around a central server (or a coordinated collection of servers). Trusted entities submit signed original images
to this server to obtain a semantic certificate that includes information about various faces detected in the image, as well as a feature representation of facial identities. This certificate is signed 
and stored as a part of the image meta-data.
Subsequently, this information can be used by an end-user, through querying the server, to verify that the image's semantic content has not been changed from the version that was submitted by the source (see Figure~\ref{F:example}). We show that this system enables verification that is robust to standard editing operations while being secure against attempts to spoof semantic categories.
All code is openly available at \url{https://github.com/manavkulshrestha/Imgprov/}.

\smallskip
\noindent{\bf Related Work }
While tampering with images and videos in order to distort scenes has been an issue for decades, advances in deep learning for image synthesis (e.g., Generative Adversarial Networks (GANs)), has made it significantly easier to create photo-realistic image and video synthesis (often called ``deep fakes'')~\cite{brock2018large,Dekel_2018_CVPR,karras2018progressive,Zhan_2019_CVPR,Zollhofer18,Deepfakes18}.

The principal approaches for countering the deleterious impact of deep fakes fall under the broad umbrella of \emph{media forensics and image forgery detection}, which aim to determine whether a particular entity (e.g., image or video) is benign (that is, represents benign goals, such as posting a selfie) or malicious (e.g., deep fakes).
At its core, detection of malicious instances is one of the fundamental problems in cybersecurity, with spam and malware detection prototypical examples~\cite{Anderson20}.
In the context of verifying authenticity of digital media, considerable effort has been devoted specifically to detecting and identifying the real sources of social media posts~\cite{Alizadeh20,Bail20}, as well as images and video~\cite{Piva13,Rocha11}.
Image and video forensic techniques have ranged from identifying the particular device on which the image or video was captured~\cite{Sencar12}, to detecting forgery and image/video tampering~\cite{Farid09}.
Classical approaches to forgery and tampering detection are often based on the hypothesis that these are imperfect and introduce detectable inconsistencies~\cite{Carvalho13,Carvalho16,Farid09,mccloskey2018detecting,Rocha11}, or can be decomposed into a detectable series of image processing operations (scaling, rotation, etc)~\cite{Bayram06}.

As deep neural networks increasingly dominate image synthesis, machine (especially, deep) learning is increasingly prevalent in forensics and detection~\cite{Bappy17,Bayar16,mo2018fake,roessler2019faceforensicspp,Huh2018,Zhou2018}.
Commonly, these frame deep fake detection as a binary classification problem, akin to malware and spam detection, with the goal of distinguishing ``fake'' images from real.
The central vulnerability with all deep fake detection techniques is that the particular insights in these can, in turn, be used to improve synthesis.
This is particularly true for GANs, which can make use of detectors directly in the learning algorithm to improve synthesis quality.
Consequently, the resulting arms race appears to strongly favor the adversary.

Our approach takes an orthogonal route,  leveraging cryptographic signing together with higher-level semantics we wish to preserve in order to provide reliable and flexible image provenance.
This idea is related to  \emph{semantic hashing}~\cite{Cao16,Chaidaroon17,Salakhutdinov09}, which is designed to speed up retrieval of textual and visual information.
The central idea in semantic hashing is to construct low-dimensional representations of inputs (documents, images) that can map similar inputs to nearby addresses in memory, significantly speeding up retrieval.
Since semantic hashing techniques are primarily designed for information retrieval, they are not necessarily secure, and cannot guarantee provenance.
For example, a common implementation of semantic hashing is through deep neural network input embedding, which is vulnerable to attacks such as face swapping (since faces are still in an image) and adversarial examples (since the entire image is used as an input).
Robust semantic hashing does not address this issue, since robustness in this context is  to non-adversarial noise rather than adversarial tampering~\cite{Yang15}.
In contrast, our approach uses secure cryptographic hashing for specific semantic categories in images, enabling semantic provenance and reducing the scope and effectiveness of black-box adversarial perturbation attacks.

\section{System Model and Problem Statement}

We consider an image sharing ecosystem comprised of three parties: 1) the \emph{image contributor}, 2) the \emph{image modifier} (possibly malicious), and 3) the \emph{image consumer}.
The image contributor is the device that captures that original image.
The resulting digital image is then shared with an \emph{image modifier} who edits and then posts it online for public consumption (this may be the photographer, a third party, etc).
Finally, the \emph{image consumer} is a lay person who sees the image posted online, receives it in an email, etc.
Note that this model is more general than it may at first seem.
For example, the contributor and modifier of the image may be the same person---for example, the photographer may first crop the image before making it available to others.
Similarly, there may well be many modifiers of the image, modifying it in parallel or sequentially.
And, of course, there may be no modifications at all.
Still, what ultimately matters in our context is the fact that the workflow involves first the creation of the image, then, possibly, this image is modified, and, finally, seen by the image consumer in this modified form.
The key issue of interest is that \emph{image modifications may be malicious}, for example, \emph{deep fakes that maliciously replace some of the semantic content of the original image in order to change the semantics of the scene}.

We aim to solve the following problem: given the image life cycle model above, we wish to ensure that we can preserve the provenance of the relevant semantic information in the original image, raising an alarm when modifications to the image are suspected of violating the integrity of such semantic information.
At the same time, we wish to ensure that we limit false positives, that is, we do not raise an alarm very often when it is not warranted (modifications are benign and preserve the relevant semantics).
Since a major threat of image integrity pertains to the identities of individuals captured in the scene (for example, maliciously placing individuals in compromising scenes), as well as their relation to the scene, we focus on two types of semantic information: 
the identities of individuals and their relative positions in the scene, as well as whether the scene was indoors or outdoors, as the semantic information we aim to verify.

\begin{figure}[t]
    \centering
    \includegraphics[scale=0.13]{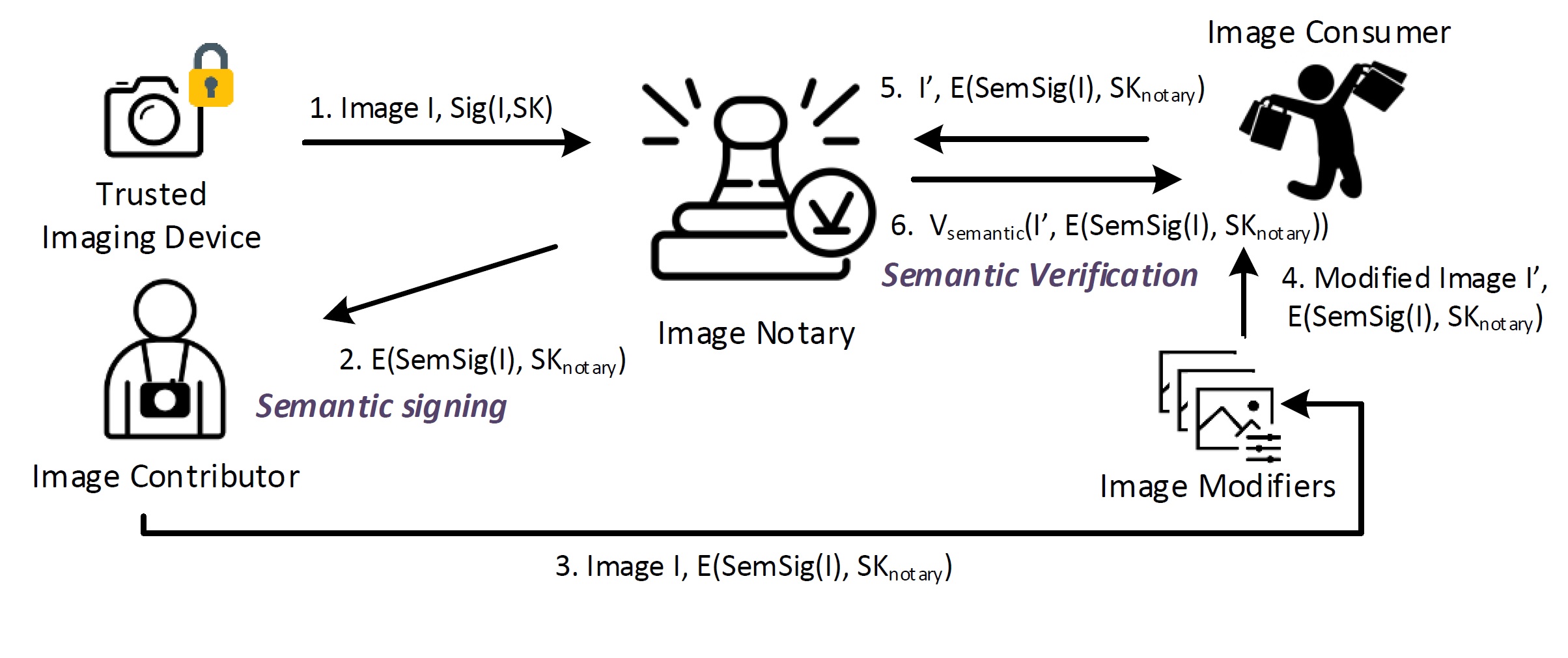}
    \caption{An image life cycle within the \textsc{PROVES} architecture.}
    \label{fig:sys-overview}
\end{figure}

\section{Proposed Solution: The PROVES System}

We propose \textsc{PROVES} (image PROVEnance using Semantic signatures), a novel system architecture that aims to establish provenance of semantic information in images.

\subsection{PROVES System Architecture}

To the three parties described in our system model, we now add a fourth: an \emph{image notary}, which is a server or a collection of servers that play a central role in both signing and verification of the semantic information in images.
The proposed \textsc{PROVES} system involves the addition of this image notary, and an architecture of interactions among the image contributor, 
the image consumer (more precisely, a web browser which displays the image to the image consumer), and the image notary (see Figure~\ref{fig:sys-overview}).

The \textsc{PROVES} architecture is implemented as a smartphone app on the image contributor device, an HTTPS web service on the notary, and a browser plugin.
The image notary exposes its signature and verification APIs, relying on SSL certificates and an established certificate authority 
to protect calls to this API and certify the responses as genuine. 
This authority will maintain a database of public keys of trusted image sources, which will typically include trusted devices for signing at the time of capture, but may also include reputed individuals and organizations that are deemed as trustworthy. It will allow for these keys to be revoked at any time, at which time the server will refuse to verify any images whose provenance was established based on information from the corresponding source prior to the revocation.
Next, we overview the workflow of the \textsc{PROVES} system.
Subsequently, we provide the details of two key steps: semantic signing and semantic verification.

In the \textsc{PROVES} architecture,
when the image is captured by the image contributor device, 
the device first signs the raw pixels of the captured image, and then sends the image and its signature to the notary with a request for a \emph{semantic signature}.
The notary first verifies the integrity of the contributor device's signature, and then returns the  signature (signed by the notary's private key) which contains an encoding of the image semantics, the ID of the device, and the signature date (according to the notary's clock).

The nature of the semantic encoding by the notary will depend on the nature of the information sought to be certified. 
In this work, we seek to be able to certify the identities and relative locations of human faces in the image, as well as whether the image was taken indoors or outdoors.
Suppose that we have a way to encode a given identity (face) as a feature vector.
For example, we can learn a deep neural network model for face recognition to obtain such a representation.
Specifically, consider an input image $x$ which contains $K$ faces.
By running a face detection algorithm $g$ (say, a neural network for face detection) and cropping appropriately, we can obtain a collection of corresponding face images, $\{x^1,\ldots,x^K\} = g(x)$.
Next, for each face image $x^k$, we can obtain an associated feature vector $y^k = f(x^k)$, where $f$ is the representation model (trained to recognize faces, say).
The server will use a face detection 
Our semantic encoding will then include: (a) the original image size and bounding boxes of faces detected in the image with respect to this image size, and (b) a feature vector encoding identity for each detected face generated by $f$.
The case of classifying whether the image was taken indoors or outdoors, we can make use of standard image classification techniques, and simply sign the binary label $z = h(x)$ obtained from the original image $x$, where $h(x)$ is the image classifier.
The complete details of the signing process are provided in Section~\ref{S:sign}.

When the device receives the semantic signature from the notary, it attaches it to the EXIF meta-data of the image file. At this point, the image may be edited---cropped, scaled, tone-mapped, etc.---retaining the original signature in its EXIF tags. 
At this point, the image is shared with others who may edit it.

Typically, we expect edits to the image to be benign.
However, we allow the possibility that they are malicious, and attempt to modify the image to qualitatively alter its semantics, for example, with respect to the identities of the people whose faces appear in it, or about where the photograph was taken.
The verification process below aims to ensure that any such malicious changes to the faces or background do not go undetected.
Specifically, when an image consumer (a web browser) downloads an image that contains a semantic signature, it proceeds to send it to the notary to verify that the image has not been semantically modified.
On receiving such a request, the notary will decrypt it, and perform the following verification process:
\begin{enumerate}[topsep=0pt,itemsep=-1ex]
  \item It will ensure that the ID of the image source has not been revoked. It will refuse to certify an image should the signing date be after the effective revocation date (which is the date that the source's private key is believed to have been compromised). For images signed before the signing date, it will continue with the certification process but also include a warning.
  \item  It will then detect faces in the given image, match the identity vectors of the detected faces to the encoded identity vectors in the signature.
  \item It will then verify the spatial configuration of the corresponding detected face bounding box locations to those in the original image, allowing for scaling, translation, and cropping, and accounting from small deviations in bounding box locations from the face detector. When verifying this configuration in the case of missing faces (i.e., those that were in the image when signed but are not present in the edited image), it will check if the face has been cropped out (i.e., its location would have been outside the image extents) or removed (i.e., the face should have appeared in the image but was artificially removed). It will return a warning for the former case, while the latter will trigger an alert.
  \item Finally, it will use an image classifier to predict whether the image was taken indoors or outdoors.  This prediction will be compared to the signed label embedded in the image, resulting in three outcomes: 1) successful verification, if the labels match, 2) failed verification if the labels differ, and 3) a warning if the confidence of the predicted label is below a threshold (see Section~\ref{S:verify} for details).
\end{enumerate}
The notary will return the information about a) which faces have been successfully verified, b) which faces, or regions of the image that should contain faces, failed verification, c) how many faces in the original image have been cropped out, and d) whether the image setting (indoors or outdoors) has been successfully verified.
The full details of the verification process are provided in Section~\ref{S:verify}.


\subsection{Semantic Signing of Images}
\label{S:sign}

At initialization, the device $c$ belonging to the contributor will generate a key pair $(PK_{c},SK_{c})$ and perform a basic handshake with the notary in order to register itself as a client with the server and provide the server with the device's public key. 
The notary will respond with its own public key. 
We assume we can reuse the existing public key infrastructure for this, or to have the notary provide the necessary service.
Furthermore, we assume that imaging device $c$ is trusted.
For example, it may be equipped with a trusted execution environment that can protect the device and notary keys from external malicious entities as well as from the device owner, even when the device software stack is compromised, similar to the common protection of the device endorsement key. 

On capturing an image $x$, the device will use its private key to sign the SHA256 hash of $x$ (i.e., sign the raw image pixels) using the Elliptic Curve Digital Signature Algorithm (ECDSA), which yields a signature $Sig_{x} = \mathcal{S} (SK_{c},x)$. 
This signature is nonforgeable even by the imaging device owner. 
$Sig_{x}$ is then sent to the notary with a request for a corresponding \emph{semantic signature} of the image $x$.

Upon receiving $Sig_{x}$, the notary will first 
use the device's public key $(PK_{c})$ to ensure that the payload was not tampered with. 
Once this message is verified, the notary first uses $g(x)$ to obtain the collection of face bounding boxes $\{x^1,\ldots,x^K\}$ embedded in the image (more precisely, the portion of the image after cropping everything outside this bounding box), and then obtains the corresponding feature vector representation $y^k = f(x^k)$ for each of these, obtained using its representation function $f$ (e.g.,, a deep neural network).
Similarly, the server computes the label $z = h(x)$ (indoor vs.~outdoor image).
Next, the server signs both each feature vector $y^k$, as well as the coordinates of each corner of the bounding boxes $x^k$, and the label $z$ to obtain the semantic signature $SemSig_{x} = \mathcal{S}_{semantic}(x, g, f, h,  c)$, again using SHA256 hashes and the notary's private key, with ECDSA.
As a final step, the notary confirms that the resulting signature can indeed be reliably verified by ensuring that the verification step $\mathcal{V}(Sig_{x}, x, PK_{c})$ (see below for details on this step) successfully passes (there are a number of possible reasons it may fail; for example, the image compression algorithm for JPEG is lossy, and decompression may cause failure in verifying the similarity of feature vectors corresponding to the same face).
If this succeeds, the resulting feature vectors, bounding boxes, and their signatures, are sent back to the device $c$.
The device then embeds these in the image meta-data, and shares the image with others.




\subsection{Semantic Verification of Images}
\label{S:verify}

Now, we describe how an image is verified.
We implemented the client-side part of verification as a browser extension prototype.
To describe the process of verification, suppose that $x$ denotes the original image that was created and signed, and $x'$ is the modified image that the web browser is about to display.
When an image is retrieved from the web server and detected to contain a semantic signature $<x', SemSig_{x}>$ as meta-data, the browser extension sends this image to the notary for validation.

The notary first (also) checks whether the image $x'$ has the semantic signature in its meta-data.
If it doesn't exist, the notary returns a response to this effect (no data exists that can be verified).
If the signature $<x', SemSig_{x}>$ is found, the server attempts to verify each semantic class.
Next, we describe the precise verification process for the two semantic classes we consider: identity (face) verification, whether the photo was taken indoors or outdoors.

\subsubsection{Face Verification}

First, the notary uses its public key to extract the feature vectors $\{y^1,\ldots,y^K\}$ and bounding box coordinates $\{p^1,\ldots,p^K\}$ of faces identified in the original image $x$ during signing.
Next, the notary detects and extracts faces from the image $x'$ being verified, obtaining $\{x'^{1},\ldots,x'^{L}\} = g(x')$, and then the corresponding features and bounding box coordinates for each, $y'^{l}$ and $p'^{l}$.
At this point, we need to verify two things: 1) the correspondence between faces detected in $x'$ and those detected in $x$, and 2) that the faces in $x'$ preserve the relative position of those in $x$.
To do this, we must account for a variety of benign image modifications.
We specifically consider small rotation, translation, scaling, and down/up-sampling the image as benign transformations.
Cropping is treated with more nuance: we identify all faces that are outside the boundaries of image $x'$, and raise an alert that the corresponding number of faces have been cropped out.
However, if a face is partially cropped, we first attempt to verify it, and only treat it as cropped out if verification fails.
Finally, if faces have been swapped or significantly tampered with, we treat the resulting modifications as adversarial and raise an alert to this effect.
Next, we describe the details of how we verify the faces  in $x'$.

First, we identify a collection of pairs of locations $\{(p^k,p'^l)\}$, where each pair is determined by the similarity between corresponding feature vectors.
Specifically, for each face $x^k$ in the original image, we identify its counterpart $x'^l$ in the new image as that which maximizes cosine similarity between the two feature vectors.
Thus, given a feature vector $y^k$, let $l^* = \argmax_l \cos(y^k,y'^l)$.
Then if $\cos(y^k,y'^l) \ge \theta$ for a predefined similarity threshold (determined empirically in a way that limits false positive rate), we add the corresponding pair $(p^k,p'^{l^*})$ to the list.
This gives us a list of \emph{seed} matches.
Note that we expect this matching step to be imperfect, since face detection methods are imperfect and often unstable.
The next set of steps ensures robustness of verification to such instability.

Our second step is to deal with translation, scaling, and rotations of images.
We estimate the parameters of translation and scaling using the seed list of bounding box pairs from the original image $x$ and the current image $x'$.
Consider a pair $(p^k,p'^{l})$, where $p^k_{x}$ is the $x$-coordinate of $p^k$ and $p^k_y$ is its $y$-coordinate, and similarly for $p'^l$.
Then we can describe these transformations for this pair as
$p'^l_x = s \cdot p^k_x + \alpha$ and $p'^l_y = s \cdot p^k_y + \beta$, which we can write in matrix-vector notation as $P^k w = p'^l$, where $w = [s,\alpha,\beta]^T$.
Next, we simply vertically stack all the $P^k$ matrices and $p'^l$ vectors to obtain a combined $P$ matrix and $p'$ vector, giving us a system of linear equations $Pw = p'$.
We now simply find the best transformation parameter fit $w^*$ which we can do by minimizing the mean squared error (MSE). That is, we solve $\min_w ||p'-Pw||^2$,
which we can do in closed form, to obtain the optimal $w^* = (P^TP)^{-1} P^Tp$.

The third step of the verification process is to  apply the learned transformation $w^*$ to each face location in the original image $x$ to obtain the expected locations of these in $x'$.
For each face $k$ in $x$, after applying transformation $w^*$, there are three possibilities: 1) it is entirely inside the bounds of $x'$, 2) it is entirely outside of $x'$ (i.e., cropped out), and 3) it is partially cropped.
If $k$ is supposed to be entirely inside $x'$, we crop the corresponding expected area of this face in $x'$ based on its \emph{original} bounding box in $x$, and then check if cosine similarity between the original feature vector $y^k$, and the feature vector extracted from its transformed and cropped version in the new image $y'^k$, is above the threshold $\theta$.
If it is, we mark this face as \emph{verified}.
If not, we mark this face as \emph{potentially tampered}, and associate with it an alert.
To account for a limited range of (discrete) translations, we repeat this step for each translation, and mark the face as verified if \emph{any} associated similarity is above $\theta$.
If $k$ is partially cropped, we attempt the verification process above.
If it succeeds, we also mark this face as \emph{verified}.
However, if it fails, we simply add it to the collection of faces that are reported as cropped.
Finally, if the face is fully outside of $x'$ post transformation, we also add it to the cropped list.
We then report the number of cropped faces.

When the verification process is complete the notary sends the semantic signature verification result $\mathcal{V}_{semantic} (SemSig_{x},x', g, f)$ back to the browser. 
This result includes a list of face bounding boxes in the image $x'$ that have been verified successfully, the list of bounding boxes that have failed verification (marked as suspicious) and should therefore trigger an alert to the user, and the number of faces detected in the original image $x$ that have been identified as cropped in $x'$.

The final step of verification takes places on the client-side.
Once verification response has been received, the interface marks bounding boxes of successfully verified faces in $x'$ by a blue box, bounding boxes of faces that failed verification in $x'$ by a red bounding box, and reports the number of cropped out faces (if any) next to the displayed image.

\subsubsection{Verifying Scene Location}

First, the notary extracts the label $z$ that documents whether the original image was taken indoors or outdoors from the semantic signature.
The simplest next step is to now use the image classifier $h(\cdot)$ to predict the label $z' = h(x')$ in the image $x'$ being verified, and then simply check whether $z' = z$.
However, we can endow the process with greater flexibility and reliability by introducing a third option where we return a warning that we do not have sufficient confidence to verify the input image.
To do this, we slightly abuse notation and redefine $h(\cdot)$ to return not the label, but the predicted probability of the reference class (say, indoors), $q$; we label the reference class 1, and the other class 0.
Next, define a threshold $\gamma$, and let $z' = 1$ if $h(x') \ge \gamma$, and $z' = 0$ if $1-h(x') > \gamma$.
If one of these two conditions is true, $z'$ is well-defined, and we perform verification as above.
If neither is true, we return a warning that we failed to verify the input due to insufficient confidence.

\subsection{Threat Model and Security Analysis}

We assume that the image notary, image contributor, and image consumers are all honest, while the image modifier may be malicious, in which case they would attempt to alter the image to change its semantic meaning. 
The assumption of an honest image notary is fundamental, as that is the essence of the proposed framework, while an honest image consumer is inconsequential for our purposes, since this framework only targets honest image consumers.
That the image contributor is honest is a strong assumption, which we can enforce in practice by being highly selective about the set of contributors whose images we would be willing to sign, and otherwise rely on the public key infrastructure to ensure that their private keys are valid.
This would rule out attacks where an image is modified and then resubmitted by a malicious party for signing (since the request of this ``image contributor" would be rejected by the server).

We assume that the adversary (malicious image modifier) has complete read and write access to the original image $x$, and the face detection algorithm $g$ used by the notary.
However, we assume that the adversary does not know $f$ or $h$ (in practice, they would not know $g$ either, but we demonstrate below that our approach is robust even if the adversary knows $g$).
Furthermore, we assume that the adversary does not have the data used to train $f$ or $h$, and can only use another dataset to train a proxy $f'$ or $h'$.

Since the notary may not control the public key infrastructure, public/private certificates of compromised devices are invalidated through explicit revocation. 
Specifically, when keying material of the contributor is compromised, we assume that the notary is immediately notified and will no longer be verifying semantic signature for any images that were signed by this contributor.
Finally, we rely on the security of standard cryptographic algorithms and libraries.

Finally, we assume that the attacker has one of the following three goals in modifying the image, all while avoiding being detected: 1) replace a face in an image with a face of another target individual, or change the background to reflect an indoor environment, when the photo was taken outdoors, and vice-versa (\emph{Replacement Attack}), 2) swap two faces in the image (\emph{Swap Attack}), and 3) remove one of the faces from the image (\emph{Removal Attack}). 
In addition, as a part of these attacks, the adversary can add adversarial noise $\delta$ to the modified face image (within its bounding box) or the background such that $\|\delta\|_\infty \le \epsilon$, where the value of $\epsilon$ is small enough to ensure that the perturbation is imperceptible (in our case, at most 4/255, at which point it becomes perceptible).



We now investigate the attack surface of the proposed system.
We assume that we can trust the signing and verification portions of PROVES, as well as the joint image capture and signing step (that is, we assume that we can trust the device, as well as the security of standard cryptographic protocols and libraries). By leveraging the PKI infrastructure, the communication will be authenticated using certificates, protecting the system from man-in-the-middle attacks. Furthermore, the requests and responses are protected using transport layer protection with nonce, such as TLS, therefore, it is resistant against replay attacks. As a result, the attack surface in our threat model are therefore restricted to adversarial modifications of the input image $x$ \emph{between the signing and verification steps}.

Consider first attacks on face detection $g(x')$, and suppose that we have replaced a particular face in $x'$ relative to $x$.
For this, suppose that $f(x)$ is reliable and robust.
There are three possibilities: 1) $g(x')$ fails to detect the new face, 2) $g(x')$ detects it correctly, and 3) $g(x')$ detects the face, but shifts its bounding box.
In the first case, our system will estimate benign modifications (translations, scaling, rotation) without including this face, and since these modifications apply to the entire image (that is, to all faces identically), \textsc{PROVES} will simply use the originally detected bounding box for this face, apply estimated translation and scaling (and rotation, if relevant), and detect that the face is an imposter by comparing the feature vector of the original face with the new face in the resulting bounding box.
If there are no other faces in the image, then the notary will simply return that it was unable to verify any faces (i.e., that initially there was 1 face, and now none can be found).
In the second case, verification will fail because the direct comparison between the original and new face will reveal the imposter.
Finally, in the third case, if the shift is large, verification will fail due to differences in feature vectors.
We evaluate the effect of small shifts in the experiments below, showing that they do not offer the adversary significant leverage.
The same set of arguments applies for face swapping.
In the case of cropping a face out of the image, verification will simply report the number of images that have been cropped out.
Finally, if the face is removed but not cropped out, the detector will fail to find it, and the same reasoning applies as in case 1) above.

The arguments above assumed that $f(x)$ allows us to reliably determine whether particular faces $x_1$ and $x_2$ belong to the same individual given their feature vectors, by checking whether $\cos(f(x_1),f(x_2)) \ge \theta$.
Since we assume that the adversary does not know $f(x)$, our experiments below consider the robustness of $f(x)$ to black-box adversarial example attacks.
Similarly, we consider adversarial robustness of $h(x)$.
\section{Evaluation}

We evaluate \textsc{PROVES} in two respects: 1) its performance with images subject to benign transformations, and 2) robustness of feature extraction $f(x)$.
In the face recognition task, we use a multi-task cascaded convolutional network~\cite{mtcnn} as the face detector $g(x)$.
We conduct experiments on two separate face extractors, namely, Inception-Resnet network (IRN)~\cite{inception-resnet} and Squeeze-and-Excitation network (SEN)~\cite{hu2018senet}. 
For determining whether two face feature vectors $y$ and $y'$ belong to the same individual, we use the threshold on cosine similarity $\theta = 0.7$.
For detecting faces in the original image $x$ when it is signed, we filter out all bounding boxes whose size is below the fraction 0.005 of the size of the full image.
However, at the time of verification, we do not filter out any detected faces.
We used the CelebFaces dataset~\cite{liu2015faceattributes} to train the feature extractor, with data augmentation using color, brightness, and contrast modifications by factors of 80\% and 120\%.
We used the Images of Groups (IoG) dataset~\cite{gallagher_cvpr_09_groups} to test that full \textsc{PROVES} pipeline in this context.
In the indoor-outdoor classification task, 
we train on the Places365-Standard dataset~\cite{zhou2017places}, which contains images from 365 different scenes, but we only use indoor/outdoor meta-class as the binary ground truth labels for training and testing. The model $h(x)$ we use for classification is WideResNet18~\cite{wideresnet}. 


\subsection{The Face Recognition Task}

\noindent{\bf Effectiveness with Benign Transformations }
Our evaluation of effectiveness investigates how successful the verification process is when the image $x$ undergoes a series of benign transformations involving scaling, translation, and (small) rotation.
Specifically, for each image, we first scale it uniformly at random in the range $[0.85,1.15]$, then translate it uniformly at random in the range of $[-0.15,0.15]$ of each image dimension (the negative means negative translation, with all translation being computed as a fraction of the corresponding dimension), with the portion of the image that goes out of bounds of the original being cropped.
Next, we perform a uniform random rotation of $[-5^\circ,+5^\circ]$.
In addition, we perform a sequence of contrast/brightness/color balance adjustments. The factor that controls the strength of each kind of adjustment is chosen uniformly at random in the interval $[0.85,1.15]$.
While a factor of 1.0 gives the original image for all adjustments, a factor of 0.0 gives a solid grey image for contrast adjustment, a black image for brightness adjustment, and a black and white image for color balance adjustment. 

Table \ref{table_angle} shows the results of applying the \textsc{PROVES} pipeline to 754  images from the IoG dataset, consisting of a total of 3705 faces.
We observe an extremely low false positive rate of 0.1\% when using IRN, and 0.4\%-0.8\% using SEN, with even partially cropped faces successfully verified in most cases.
\begin{table}[h]
	\centering
	\caption{Effectiveness of \textsc{PROVES}, showing the \% of successfully verified faces and verified partially cropped faces. 
	}
	\begin{small}
\begin{tabular}{||c||c|c|c|c||}
		\hline
		\multirow{2}{*}{\textbf{Rotation Angle}} & \multicolumn{2}{c|}{IRN} & \multicolumn{2}{c||}{SEN} \\\cline{2-5}
		
		& 0\textdegree & 5\textdegree & 0\textdegree & 5\textdegree\\\hline\hline
		{Verified}                         &  94\%  & 92.9\%  & 96.9\% & 96.1\% \\ \hline
		{Verified Partial Faces}         &   5.9\%  &  7\% &   2.7\%  &  3.1\% \\ \hline
		{Fail to verify}                   &   0.1\%   &   0.1\% &   0.4\%   &   0.8\% \\ \hline
	\end{tabular}
	\end{small}
	\label{table_angle}
\end{table}
Furthermore, we find that both signing and verification steps are fast, taking typically under 2.5 seconds even for images with up to 10 faces.


\smallskip
\noindent{\bf Adversarial Robustness }
We now study the vulnerability of face feature extractor $f(x)$ to adversarial perturbation attacks, that is, attacks which add imperceptible noise to images in order to fool predictions~\cite{goodfellow2015explaining,DBLP:conf/iclr/MadryMSTV18}.
Since our threat model assumes that $f(x)$ is not known to the adversary directly, our analysis is focused on its robustness to black-box adversarial example attacks.
Black-box attacks commonly leverage transferability of adversarial examples~\cite{Vorobeychik18book}.
We use the following common implementation of such attacks.
First, we split a training dataset for learning $f(x)$ into two equal-sized subsets, $D_1$ and $D_2$.
We then train two versions of face feature representation model, $f_1$ (on $D_1$) and $f_2$ (on $D_2$).
Finally, we deploy $f_2$ as the \emph{true} model used by the notary, whereas the attacker uses $f_1$ to devise the input-dependent adversarial perturbations $\delta$, which are then evaluated using $f_2$.

To design adversarial perturbations $\delta$ we use the state-of-the-art PGD (projected gradient descent) attack~\cite{DBLP:conf/iclr/MadryMSTV18} using step size 0.01.
Note that our attacks are effectively \emph{targeted}: we aim to make the impersonator face be classified as the original face.
Specifically, if we fix the feature vector $y$ of the original face, the adversary aims to minimize $\cos(y,f_1(x+\delta))$ with respect to $\delta$. 
We also consider white-box PGD attacks in which we directly attack $f(x)$.

\begin{figure}[h]
\centering
\begin{tabular}{cc}
\includegraphics[width=0.8\linewidth]{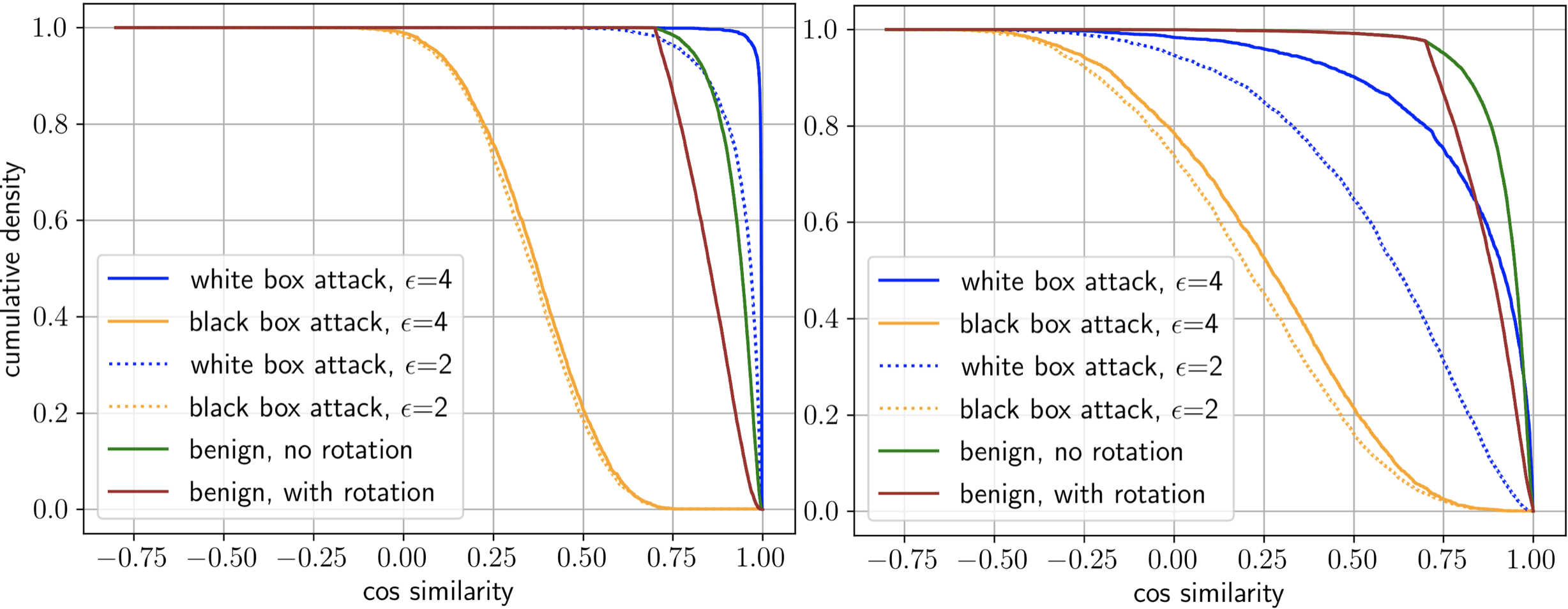} 
\end{tabular}
  \caption{Distribution ($1-$cdf) of face recognition performance for benign face image transformations (with and without rotation), as well as white-box and black-box malicious perturbations, for the IRN model.  Left: regularly trained face recognition model.  Right: adversarially trained face recognition model.
  }
\label{fig:eps4_cdf}
\end{figure}
Figure \ref{fig:eps4_cdf} (left) presents the results comparing prediction performance of $f(x)$ that uses IRN with benign transformations to black-box and white-box attacks that use $\epsilon = 2$ and 4 (out of 255); results with SEN are similar.
We can see a clear separation between the cosine similarity values of face images subject to black-box attacks compared to faces with benign transformations.
Consequently, $\theta=0.7$ correctly classifies the vast majority of faces (see above), but identifies nearly all manipulated cases.
However, note that the approach is vulnerable to white-box attacks.
While we can make our approach ($f(x)$) more robust to white-box attacks by adversarial training (see Figure \ref{fig:eps4_cdf} (right)), this comes at a significant cost to  successful verification with benign modifications (the false positive rate increases to 7\% from 0.1\%), and even at some cost to robustness to black-box attacks.

\begin{figure}[!h]
\centering
\begin{tabular}{c}
\includegraphics[width=0.8\linewidth]{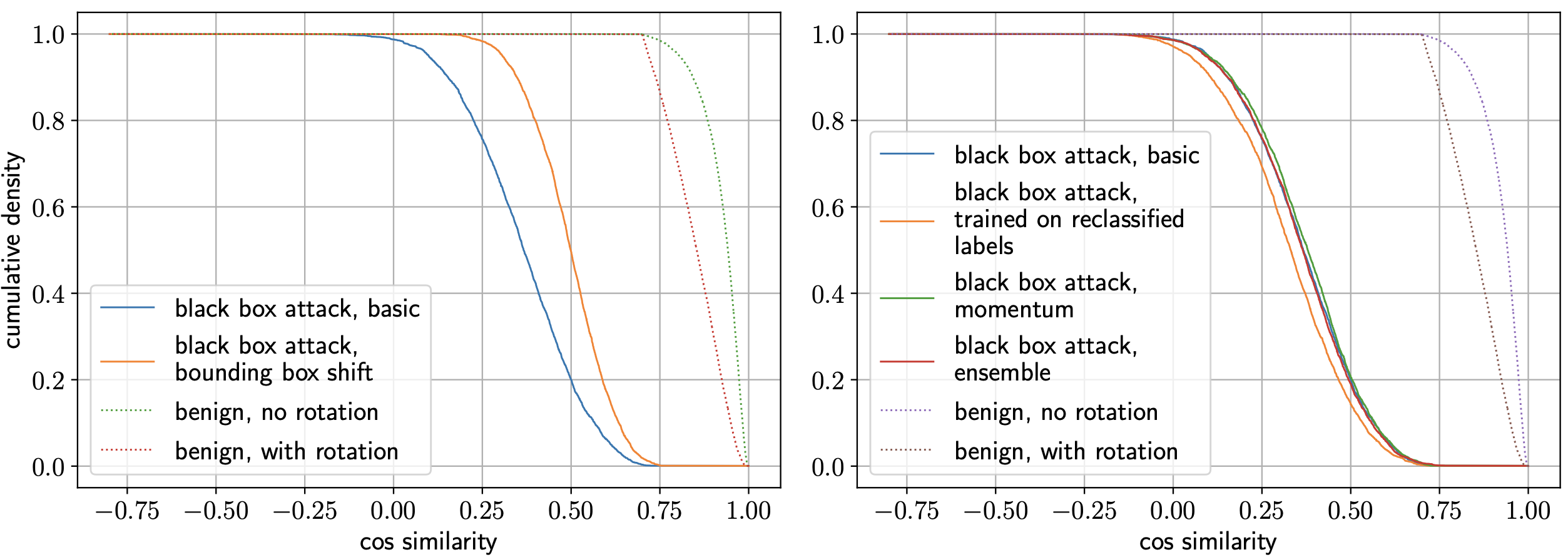}
\end{tabular}
\begin{tabular}{c}
\includegraphics[width=0.8\linewidth]{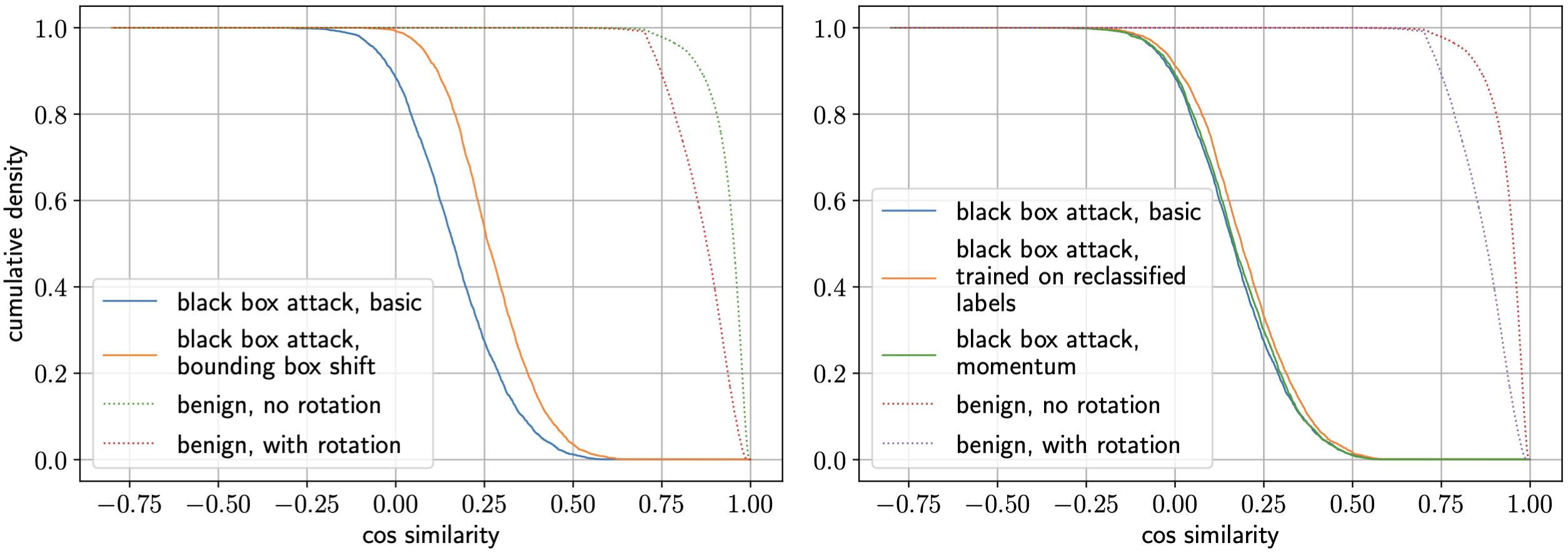}
\end{tabular}
  \caption{Different variants of black box attacks.  Top: IRN model.  Bottom: SEN model.
  }
\label{fig:faceNet_blackVariant}
\end{figure}
Next, we consider the impact of shifting the bounding box on the efficacy of black-box attacks.
We consider five positions of the bounding box: original, and 4 shifts by 10\% (up, down, right, left).
For each, the attacker designs and adversarial perturbation, and then chooses the most effective shift and associated perturbation.
As shown in Figure~\ref{fig:faceNet_blackVariant} (left), while this attack is slightly stronger than the baseline variant, there remains a clear separation between benign and adversarial modifications for both IRN (top) and SEN (bottom) models.
Finally, we consider several stronger variants of black-box attacks, 
including those using an ensemble~\cite{tramer2017ensemble}, momentum~\cite{dong2018boosting}, and which use predicted rather than original labels to learn a proxy model~\cite{Vorobeychik14}.
As shown in Figure~\ref{fig:faceNet_blackVariant}, these are not significantly more effective than the baseline black-box attack in our setting.

\subsection{Indoor vs.~Outdoor Classification}
Next, we evaluate the effectiveness of verification in the indoor-outdoor classification setting with benign and adversarial transformations.

\begin{figure}[h]
\centering
\includegraphics[width=0.85\linewidth]{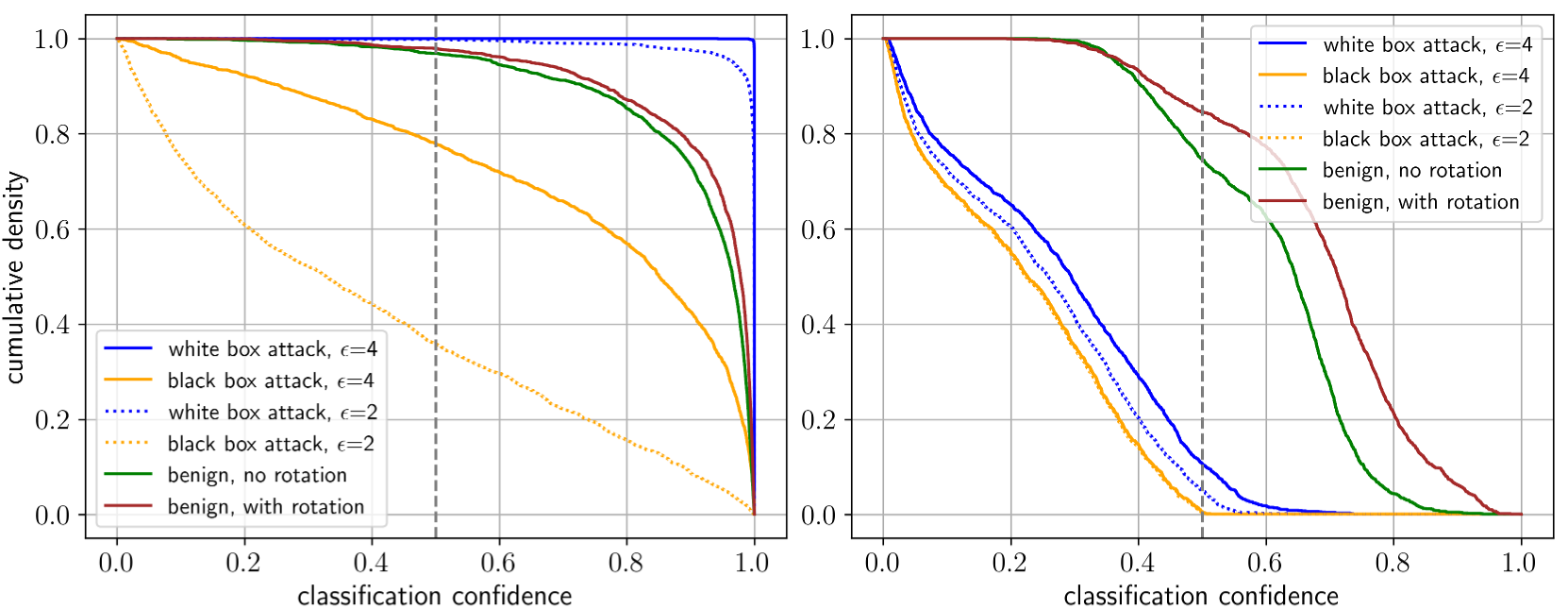}
  \caption{ Distribution (1-cdf) of indoor/outdoor classification performance under benign transformations and adversarial attacks. The vertical grey dashed line is for $\gamma=0.5$.
  }
\label{fig:io_cdf}
\end{figure}
Figure~\ref{fig:io_cdf} shows 1-cdf of classification decisions obtained as we change the decision threshold $\gamma$.
From the left figure, we can see that both black-box and white-box attacks are highly effective against the conventionally trained model.
With adversarial training, however, while accuracy decreases for benign transformations, we are able to achieve considerable separation between benign and adversarial modifications.
For example, we get get over 90\% prediction success rate with attack efficacy under 20\%.

\section{Conclusion}

Modern image synthesis techniques have given rise to powerful new deception tools in the form of deep fakes.
A typical mitigation approach is to use deep neural networks to predict which images are fake, but this approach appears to be losing an arms race, as image synthesis improves by evading detection.
We take a fundamentally different approach, using secure cryptographic signatures to sign semantic content of an image, such as faces, captured by a trusted source.
We can then reason about both benign (translation, rotation, scaling) and malicious (face swapping, adversarial noise) modifications to the semantic content to assure image provenance.
Our framework partly relies on the success of adversarially robust identification of semantic content, and we experimentally demonstrate that either black-box spoofing attacks are relatively ineffective, as is the case with face identification, or adversarial training significantly boosts robustness.
While we show that our framework can be used to preserve provenance of identities (faces) and scene location (indoors or outdoors), it can be similarly applied to arbitrary semantic categories.
Moreover, as the set of semantic categories becomes richer, our framework yields more comprehensive image provenance.

\vspace{-5pt}
\paragraph{Acknowledgments} This research was partially supported by the NSF (IIS-1905558, ECCS-2020289, CNS-1916926) and ARO (W911NF1910241).






{\small
\bibliographystyle{ieee_fullname}
\bibliography{main}
}




\end{document}